\documentclass[letterpaper, 10 pt, journal, twoside]{IEEEtran}
\pdfoutput=1
\usepackage{multirow}
\usepackage{threeparttable}
\usepackage{hyperref}
\usepackage{amssymb}
\usepackage{graphicx}
\usepackage{fancyhdr}
\usepackage{cite}
\ifCLASSINFOpdf
\else
\fi
\usepackage{amsmath}
\begin{document}
\title{Visual Attention-based Self-supervised Absolute Depth Estimation using Geometric Priors in Autonomous Driving}

% \author{Michael~Shell,~\IEEEmembership{Member,~IEEE,}
%         John~Doe,~\IEEEmembership{Fellow,~OSA,}
%         and~Jane~Doe,~\IEEEmembership{Life~Fellow,~IEEE}% <-this % stops a space
% \thanks{M. Shell was with the Department
% of Electrical and Computer Engineering, Georgia Institute of Technology, Atlanta,
% GA, 30332 USA e-mail: (see http://www.michaelshell.org/contact.html).}% <-this % stops a space
% \thanks{J. Doe and J. Doe are with Anonymous University.}% <-this % stops a space
% \thanks{Manuscript received April 19, 2005; revised August 26, 2015.}}
\author{Jie Xiang$^{1}$, Yun Wang$^{2}$, Lifeng An$^{2}$, Haiyang Liu$^{2}$, Zijun Wang$^{2}$ and Jian Liu$^{2}$%
\thanks{Manuscript received: May 1, 2022; Revised August 2, 2022; Accepted September 17, 2022.}%Use only for final RAL version
\thanks{This paper was recommended for publication by Editor Cesar Cadena Lerma upon evaluation of the Associate Editor and Reviewers' comments.
This work was supported by the National Key Research and Development Program of China (No. 2021YFB2501403). (Corresponding author: Jie Xiang.)} %Use only for final RAL version
\thanks{$^{1}$Jie Xiang is with the Institute of Microelectronics, Chinese Academy of Sciences, Beijing 100029, China, and with the School of Electronic, Electrical and Communication Engineering, University of  Chinese Academy of Sciences, Beijing 100049, China.
        {\tt\footnotesize (Email: xiangjie@ime.ac.cn).}}%
\thanks{$^{2} $Yun Wang, Lifeng An, Haiyang Liu, Zijun Wang and Jian Liu are with the Institute of Microelectronics, Chinese Academy of Sciences, Beijing 100029, China.
        {\tt\footnotesize (Email: \{wangyun, anlifeng, liuhaiyang, wangzijun, liujian\}@ime.ac.cn).}}%
\thanks{Digital Object Identifier (DOI): 10.1109/LRA.2022.3210298.}
}
\maketitle
\thispagestyle{fancy}
\lhead{\small\copyright 2022 IEEE.  Personal use of this material is permitted.  Permission from IEEE must be obtained for all other uses, in any current or future media, including reprinting/republishing this material for advertising or promotional purposes, creating new collective works, for resale or redistribution to servers or lists, or reuse of any copyrighted component of this work in other works.}

\begin{abstract}
Although existing monocular depth estimation methods have made great progress, predicting an accurate absolute depth map from a single image is still challenging due to the limited modeling capacity of networks and the scale ambiguity issue. In this paper, we introduce a fully Visual Attention-based Depth (VADepth) network, where spatial attention and channel attention are applied to all stages. By continuously extracting the dependencies of features along the spatial and channel dimensions over a long distance, VADepth network can effectively preserve important details and suppress interfering features to better perceive the scene structure for more accurate depth estimates. In addition, we utilize geometric priors to form scale constraints for scale-aware model training. Specifically, we construct a novel scale-aware loss using the distance between the camera and a plane fitted by the ground points corresponding to the pixels of the rectangular area in the bottom middle of the image. Experimental results on the KITTI dataset show that this architecture achieves the state-of-the-art performance and our method can directly output absolute depth without post-processing. Moreover, our experiments on the SeasonDepth dataset also demonstrate the robustness of our model to multiple unseen environments.
\end{abstract}

% \begin{IEEEkeywords}
% IEEE, IEEEtran, journal, \LaTeX, paper, template.
% \end{IEEEkeywords}
\begin{IEEEkeywords}
Deep learning for visual perception, computer vision for transportation, range sensing.
\end{IEEEkeywords}

% \fi
\IEEEpeerreviewmaketitle

\section{Introduction}
% for IEEE journal papers produced under \LaTeX\ using
% IEEEtran.cls version 1.8b and later.
\IEEEPARstart{M}{onocular} Depth Estimation (MDE) is a fundamental problem in the field of computer vision, which refers to predicting the corresponding depth map from a single image.  Depth Estimation has broad application prospects in autonomous driving, such as 3D object detection \cite{Park_2021_ICCV}, scene understanding \cite{schon2021mgnet}, and obstacle avoidance \cite{mancini2018j}. Compared to other ways of measuring depth \cite{Uhrig2017THREEDV, hartley2003multiple}, monocular depth estimation has unique advantages in obtaining dense depth maps at low cost. Therefore, MDE has aroused the interest of many researchers. 

In the past few years, many deep learning-based MDE methods \cite{eigen2014depth, Eigen_2015_ICCV, yin2019enforcing, lee2019big, garg2016unsupervised, godard2017unsupervised, peng2021excavating, zhou2017unsupervised, Yang2018UnsupervisedLO, yang2018lego, xue2020toward, wang2021can, klingner2020self,  johnston2020self, jung2021fine, yan2021channel, guizilini20203d, wagstaff2020self} have emerged and continual improvements \cite{Eigen_2015_ICCV, yin2019enforcing, lee2019big, garg2016unsupervised, godard2017unsupervised, peng2021excavating, zhou2017unsupervised, Yang2018UnsupervisedLO, yang2018lego, xue2020toward, wang2021can, klingner2020self,  johnston2020self, jung2021fine, yan2021channel, guizilini20203d, wagstaff2020self} have been made. These methods can be divided into supervised methods \cite{eigen2014depth, Eigen_2015_ICCV, yin2019enforcing, lee2019big} and self-supervised methods \cite{garg2016unsupervised, godard2017unsupervised, peng2021excavating, zhou2017unsupervised,  Yang2018UnsupervisedLO, yang2018lego, xue2020toward, guizilini20203d,johnston2020self, klingner2020self, wagstaff2020self, wang2021can, jung2021fine, yan2021channel}. Unlike supervised methods, self-supervised methods do not need to rely on ground truth depth for training. Furthermore, existing self-supervised MDE methods can be divided into two categories: self-supervised stereo training methods \cite{garg2016unsupervised, godard2017unsupervised, peng2021excavating} and monocular training methods \cite{zhou2017unsupervised,Yang2018UnsupervisedLO, yang2018lego,xue2020toward, guizilini20203d, johnston2020self,  klingner2020self, wagstaff2020self, wang2021can, jung2021fine, yan2021channel}. The former use stereo image pairs while the latter use monocular videos as training data. In contrast to stereo training, monocular training is a more general form of self-supervised methods and easier to obtain training data. 

However, existing self-supervised monocular training methods still have difficulties in predicting accurate metric depth. This unsatisfied performance may come from two reasons: the limited modeling power of the model and the limited supervision of loss functions. Many existing self-supervised depth estimation networks \cite{xue2020toward,  klingner2020self, wagstaff2020self, wang2021can, godard2019digging} adopt residual networks \cite{he2016deep} based on an encoder-decoder architecture \cite{ronneberger2015u}. Such a depth network shows a powerful capacity for capturing local information and representing hierarchical abstract features, but is insufficient to distinguish important details from interfering information. Due to these characteristics of the model, it is difficult for these methods to predict accurate depth estimation results on patterned surfaces or tiny structures. Thus, we introduce a fully Visual Attention-based Depth (VADepth) network. Specifically, we build a decoder based on the designed visual attention blocks and use a visual attention network followed by a dual attention module as the encoder. Benefiting from the ability to dynamically handle long-range dependencies brought by combining channel attention and spatial attention at all stages of the encoder-decoder architecture, our model not only continuously preserves important details but also effectively filters out noise.

On the other hand,  most depth networks \cite{zhou2017unsupervised,Yang2018UnsupervisedLO, yang2018lego,xue2020toward, guizilini20203d, johnston2020self,  klingner2020self, wang2021can, jung2021fine, yan2021channel} trained on monocular videos only output relative depth due to lacking scale constraints. The commonly used scale recovery method called median scaling \cite{zhou2017unsupervised} requires ground truth depth data that is unavailable at test time in many practical applications. Training a model that directly produces absolute depth in metric units can omit the scale recovery operation like median scaling. An intuitive idea to this end is to use object size \cite{sucar2017probabilistic} but it suffers from seeking ubiquitous fixed-size objects and detecting the appearance size of detected objects. A simple but effective approach is to use camera height with respect to ground surface because the ground truth camera height in autonomous driving scenarios is usually fixed and known in advance. Furthermore, we also observe that the middle bottom regions of almost all images captured in these scenarios belong to the ground surface. Considering that not all points on the ground surface are coplanar, we propose to detect all coplanar points on the ground surface from the estimated depth, based on the assumption that a small rectangular region in the middle bottom of any image is part of the ground surface. Using the height errors calculated based on detection results, we construct an absolute scale loss. Combining the scale loss with photometric reconstruction loss \cite{godard2019digging}, the visual attention-based depth network is able to output accurate absolute depth map without post-processing.

In summary, we make the contributions as follows:
\begin{itemize}
	\item A novel depth network architecture that fully utilizes the long-range perception properties of visual attention to better perceive the scene structure and predict more accurate depth maps.
	\item An absolute scale loss that leverages the geometric priors in autonomous driving scenarios to train a scale-aware depth estimation model.
	\item We conduct extensive experiments to verify the effectiveness of our network structure and loss function. Experimental results \cite{geiger2012we} show that our method outperforms previous state-of-the-art methods on the KITTI dataset and our model generalizes well on the unseen SeasonDepth dataset \cite{hu2020seasondepth} that contains multiple challenging environments without fine-tuning.
\end{itemize}

\section{RELATED WORK}

\subsection{Monocular Depth Estimation}

Early deep learning-based works \cite{eigen2014depth, Eigen_2015_ICCV} on monocular depth estimation mainly focused on supervised methods and dramatically improved the state-of-the-art performance. However, ground truth depth maps for supervised methods are hard to obtain. To avoid the heavy work of collecting ground truth depth, Garg et al. \cite{garg2016unsupervised} proposed a self-supervised method inspired by view synthesis task \cite{xie2016deep3d}, which leverages stereo image pairs to construct the reconstruction loss. Compared to stereo image pairs, monocular videos are much easier to obtain. Therefore, SfMLearner \cite{zhou2017unsupervised} leveraged monocular videos to train MDE models. Unlike stereo training approaches \cite{garg2016unsupervised, godard2017unsupervised,peng2021excavating}, monocular training methods \cite{zhou2017unsupervised, xue2020toward} need to simultaneously estimate ego-motion and depth at training time. 

Recently, monocular training methods have been extensively studied. Some works \cite{Yang2018UnsupervisedLO, yang2018lego, godard2019digging, shu2020feature, wang2021can, jung2021fine} explored to design more effective loss functions based on the photometric reconstruction loss. For example, edge-aware depth-normal consistency loss was used in \cite{Yang2018UnsupervisedLO, yang2018lego}. Godard et al. \cite{godard2019digging} proposed a per-pixel minimum reprojection loss to solve the occlusion problem, a multi-scale loss for alleviating the local gradient problem, and the auto-masking method to filter out stationary pixels. Based on Mondepth2 \cite{godard2019digging},  the use of feature-metric loss \cite{shu2020feature}, temporal geometric consistency \cite{wang2021can} and semantic-depth consistency \cite{jung2021fine} further improved monocular depth estimation accuracy. Our method also follows the general pipeline of Monodepth2 \cite{godard2019digging}. 

Besides, some works \cite{guizilini20203d, lyu2021hr} proposed more powerful convolutional networks to enhance the accuracy of depth estimation. Beyond ResNet-based U-Net, Packnet-sfm \cite{guizilini20203d} leveraged 3D convolutions to improve representational capacity of the network. HR-Depth \cite{lyu2021hr} introduced more dense skip-connections to better fuse multi-scale feature maps. Because of the improvement of loss functions and network architectures, the performance of self-supervised monocular depth estimation has been improved a lot. However, these methods still face challenges in predicting metrically accurate depth due to the scale ambiguity issue and lacking the ability to distinguish important features from noise.

\subsection{Scale Ambiguity}
Because of lacking scale constraints, monocular vision suffers from the scale ambiguity issue. In order to produce absolute depth at test time, some MDE works \cite{zhou2017unsupervised,Yang2018UnsupervisedLO, yang2018lego,xue2020toward, guizilini20203d, johnston2020self,  klingner2020self, wang2021can, jung2021fine, yan2021channel} leveraged post-processing for scale recovery. The typical post-processing technique of median scaling proposed in \cite{zhou2017unsupervised} scales the relative depth map by the ratio of the median of the ground truth depth values to the median of the estimated depth values. Median scaling has been adopted by many subsequent works \cite{guizilini20203d, johnston2020self,  klingner2020self, wang2021can, jung2021fine, yan2021channel,  godard2019digging, shu2020feature}. Nevertheless, ground truth depth maps required by median scaling are not always available at test time in many cases. As the use of camera height in classical monocular visual odometry \cite{wang2018monocular}, DNet \cite{xue2020toward} leveraged a dense geometrical constraints module to determine the scale factor by the estimated camera height from every ground point. 

Compared to post-processing techniques, training a scale-aware model is more attractive due to its simplicity of testing. To learn metrically accurate depth estimation,  \cite{roussel2019monocular} pretrained depth network on stereo data of one dataset and then fine-tuned the model on monocular videos of another dataset to preserve the absolute scale, and PackNet-sfm \cite{guizilini20203d} leveraged the camera's velocity and \cite{bartoccioni2021lidartouch} utilized 4-beam Lidar data to construct scale-aware loss. All of these works require additional sensor data for training. In order to avoid using other sensor data, \cite{wagstaff2020self} proposed a camera height-based loss function to enforce metrically-scaled depth during training. Similar to \cite{wagstaff2020self}, we also leverage camera height to construct scale constraints. However, our method directly determines the pixels corresponding to the ground plane based on predicted depth instead of relying on an extra pretrained ground plane segmentation model as in \cite{wagstaff2020self}. Thus, we simplify the training procedure and improve depth estimation accuracy by mitigating the effects of the limited accuracy of the ground plane segmentation model and the uneven ground surface.

\subsection{Attention Mechanism}

Attention mechanism is a dynamic process that diverts attention to important features \cite{Guo2022}. The pioneering work of visual attention is the spatial attention network called RAM \cite{mnih2014recurrent}. After that, more attention mechanisms \cite{Hu_2018_CVPR, Woo_2018_ECCV, Fu_2019_CVPR, dosovitskiy2021an} were proposed, such as self-attention \cite{vaswani2017attention} and channel attention \cite{Hu_2018_CVPR}. Nowadays, attention mechanisms have been applied in many visual tasks \cite{mnih2014recurrent, Hu_2018_CVPR, Woo_2018_ECCV, Fu_2019_CVPR, dosovitskiy2021an, ranftl2021vision, guo2022visual}. As for monocular depth estimation, the attention-based network has been applied not only to supervised methods \cite{chen2021attention, Lee_2022_WACV} but also to self-supervised methods \cite{johnston2020self, jung2021fine, yan2021channel}. In \cite{johnston2020self}, an attention module was used to better perceive contextual information. FSRE-Depth \cite{jung2021fine} utilized an attention module to fuse depth features and semantic features. CADepth-Net \cite{yan2021channel} applied channel attention for structure perception as well as skip-connection. However, these networks \cite{chen2021attention, Lee_2022_WACV}, \cite{johnston2020self, jung2021fine, yan2021channel} just added several attention modules to the original network, and the body of the depth network is still the usual localized convolutional network. Therefore, we introduce the visual attention mechanism throughout the processing of the network, so that our model can continuously and adaptively extract important features and remove noise interference at all stages for more accurate depth predictions.

\section{METHOD}

In this section, we first introduce the basic paradigm for self-supervised monocular depth estimation. Then, we describe the details of our VADepth network. Finally, we introduce a simple method for detecting coplanar points and construct a scale loss based on the planar geometric prior. 

\subsection{Self-Supervised Monocular Depth Estimation} During the self-supervised MDE model training, the model is optimized by minimizing the difference between the target image and the image synthesized by the source image and the predicted target depth map. In the training process using monocular videos, the source frame is chosen from the same monocular image sequence as the target frame. Following \cite{zhou2017unsupervised}, the source frame $I_{s}$ is an adjacent frame to the target frame $I_{t}$, i.e. $s = t\pm1$. Since the relative pose between two adjacent frames is unknown, we need to estimate not only the depth but also the ego-motion during training. The depth network takes an RGB image $I_t$ as input and outputs a depth map $D_t$. Simultaneously, the pose network predicts the relative pose $T_{t \rightarrow s}$ between $I_{s}$ and $I_t$. Let ${K}$ denote the known camera intrinsic matrix. Then we can project the target pixel coordinates $(u, v)$ to the source view and obtain the 2D coordinates of the projected depths $D_t(u, v)$ in $I_{s}$ as follows:
\begin{equation} \label{ps}
(u_s, v_s) = KT_{t \rightarrow s}P_t(u, v),
\end{equation}

\begin{equation} \label{pt}
\text{and }P_t(u, v) = D_t(u, v)K^{-1}(u, v, 1)^T.
\end{equation}
Here, $(u, v)$ and $(u_s, v_s)$ are the projected coordinates in $I_t$ and $I_s$ of the same 3D points $P_t(u, v)$, respectively. For simplify of notation, the conversions between inhomogeneous  and homogeneous coordinates are omitted in (\ref{ps}). Because the calculated coordinates $(u_s, v_s)$ are not always integer values, we use bilinear interpolation to compute pixel values $I_s(u_s, v_s)$ as in \cite{jaderberg2015spatial}. Thus, we can synthesize the target image: 
\begin{equation} \label{ist}
I_{s \rightarrow t}(u, v) = I_s\bigl<(u_s, v_s)\bigr>, 
\end{equation}
where $\bigl<\cdot\bigr>$ denotes bilinear sampling operation. Following \cite{godard2017unsupervised}, we combine the L1 loss and structural similarity index measure (SSIM) \cite{wang2004image} to formulate the photometric error:
\begin{equation} \label{lpe}
PE(I_a, I_{b}) = \alpha ||I_a-I_{b}||_1 + (1-\alpha)\frac{1-SSIM(I_a, I_{b})}{2} . 
\end{equation}
To avoid the false photometric error of occluded pixels in the source frame, we apply per-pixel minimum photometric loss \cite{godard2019digging}, i.e.
\begin{equation} \label{lp}
L_{ph} = \min_{s\in\{t-1, t+1\}} PE(I_t, I_{s \rightarrow t}) . 
\end{equation}
Besides, we adopt the auto-masking method \cite{godard2019digging} to filter out pixels that do not change the appearance between $I_s$ and $I_t$. The binary mask $\mu$ is computed as
\begin{equation} \label{mu}
\mu = [\min_{s} PE(I_t, I_{s \rightarrow t}) < \min_{s} PE(I_t, I_{s})],
\end{equation}
where $[\cdot]$ is the Inverse bracket. Like \cite{godard2017unsupervised}, we also use the edge-aware smoothness loss:
\begin{equation} \label{lsm}
L_{sm} = |\partial_ud_t^*|e^{-|\partial_uI_t|} + |\partial_vd_t^*|e^{-|\partial_vI_t|} , 
\end{equation}
where $d_t^*=d_t/\overline{d_t}$ is the mean-normalized inverse depth. Considering the gradient locality of the bilinear sampler, we predict multi-scale depth maps in the decoder and compute the individual losses at each scale following \cite{godard2019digging}. Thus, the final baseline loss  with a hyperparameter $\lambda_{sm}$ is defined as 
\begin{equation} \label{lbase}
L_{baseline} = \frac{1}{S}\sum_{i=0}^{S-1}\mu L_{ph} + \lambda_{sm}L_{sm}, 
\end{equation} 
where $S$ refers to the number of multi-scale depth maps.

\begin{figure}[t]
	\centering
	\includegraphics[scale=1.0]{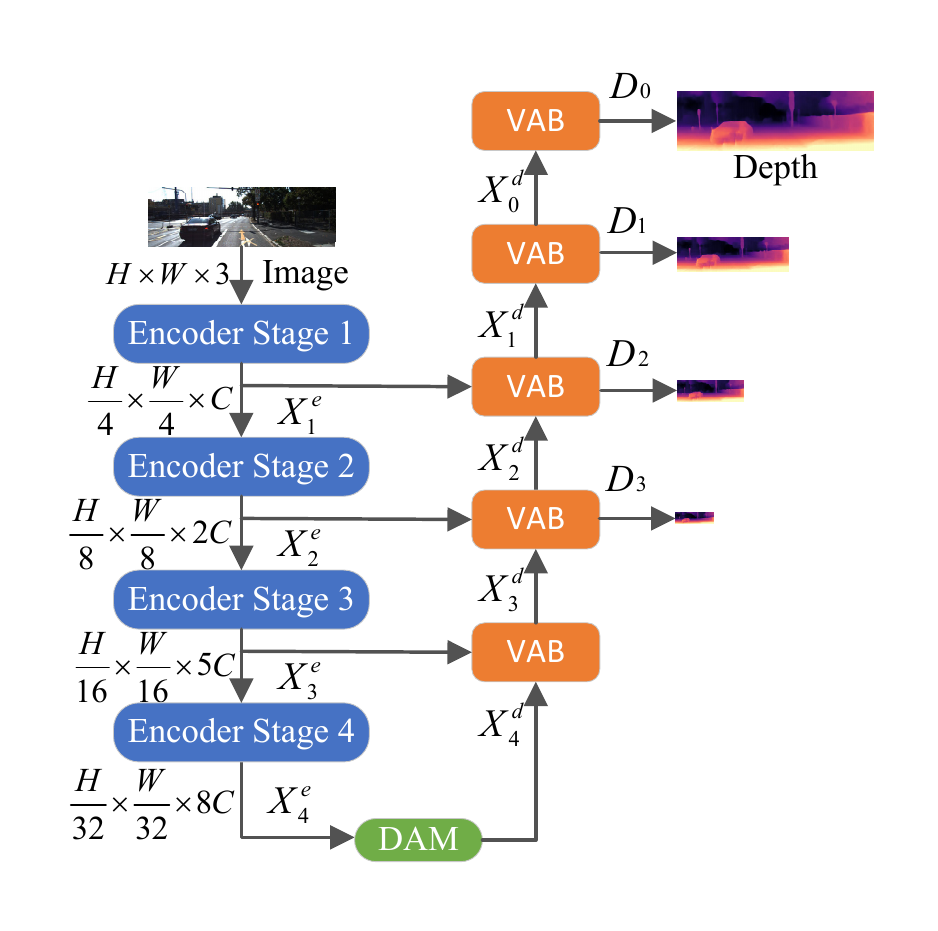}
	\caption{Network structure of VADepth. The encoder extracts feature maps $X_i^e$ at different resolutions. The Dual Attention Module (DAM) takes the last level feature map $X_4^e$ as input and emphasizes the important features to produce the initial input of the decoder. The decoder has five successive Visual Attention Blocks (VABs) and outputs multi-scale depth maps for the top four VABs.}
	\label{fig_VADepth}
\end{figure}
\subsection{VADepth Network Architecture}

\begin{figure}[h]
	\centering
	\includegraphics[scale=0.9]{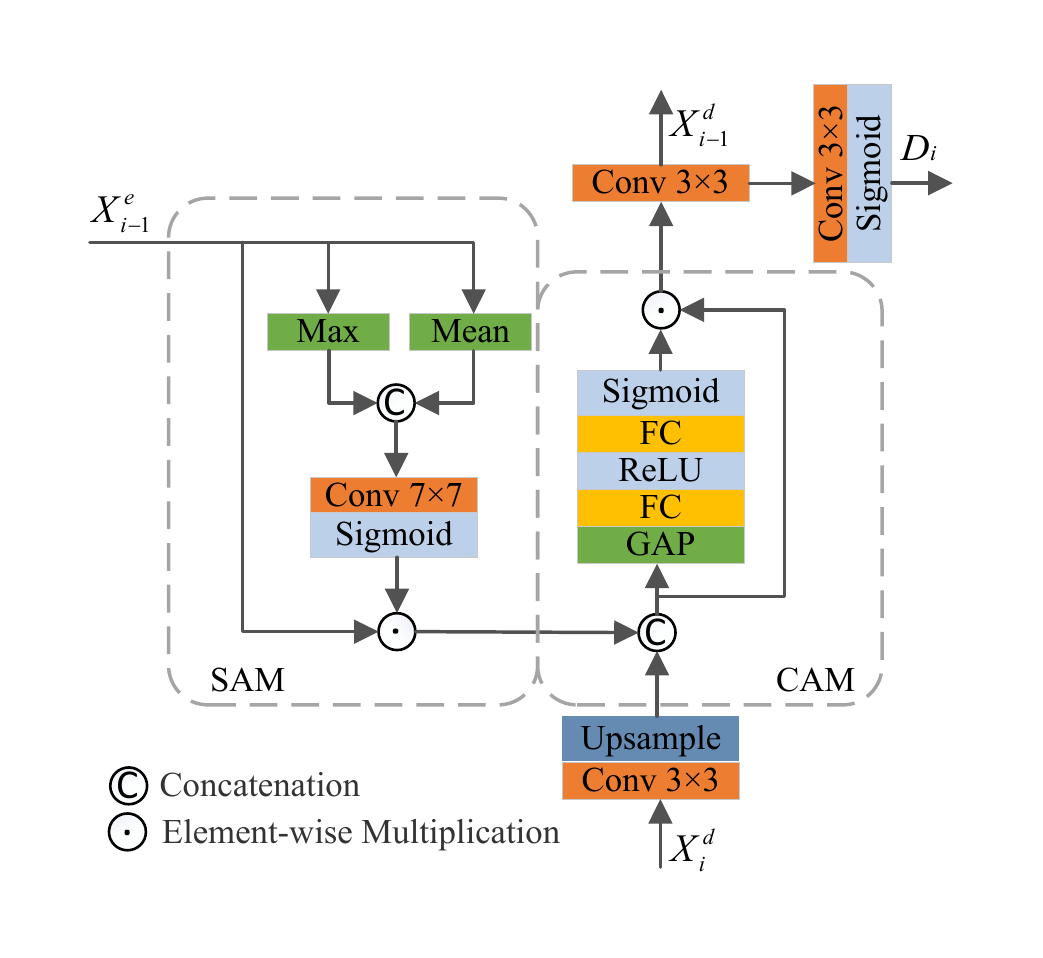}
	\caption{Network details of Visual Attention Block (VAB). A Spatial Attention Module (SAM) and a Channel Attention Module (CAM) are used in the VAB.} 
	\label{fig_VAB}
\end{figure}

As shown in Fig. \ref{fig_VADepth}, VADepth adopts the improved  U-shape encoder-decoder architecture, which takes a single RGB image as input and outputs multi-scale depth maps. Unlike U-net architecture \cite{ronneberger2015u}, the encoder and decoder in VADepth are not completely symmetrical, and an additional Dual Attention Module (DAM) head (see details in \cite{Fu_2019_CVPR}) is used to connect the encoder and the decoder. 

For the encoder, we employ a visual attention network (VAN) \cite{guo2022visual} to extract multi-scale feature maps $X^e_i, i=1, 2, 3, 4$. A VAN has four stages, where spatial adaptability and channel adaptability are efficiently implemented by the large kernel attention. 

Then the internal features of the encoder are fed into the decoder with skip connections and the lowest resolution features are taken as the input of the DAM. The DAM adaptively selects the discriminative features as the initial input of the decoder by learning spatial and channel interdependencies of features in parallel. 

As for the decoder, we design a novel Visual Attention Block (VAB). There are five successive VABs in the decoder. As illustrated in Fig. \ref{fig_VAB},  the $i\mbox{-}th$ VAB has two inputs, the internal features $X^e_{i-1}$ from the encoder and the features $X^d_i$ from the previous VAB. $X^e_{i-a}$ is fed into a Spatial Attention Module (SAM), which emphasizes the important regions and suppresses the unimportant or noisy regions as in \cite{Woo_2018_ECCV}. In the SAM, the initial spatial attention maps are formed by max pooling and average pooling along the channel axis. Then the concatenated initial attention maps are passed through a convolutional layer with a sigmoid function to obtain the final spatial attention map. The output of the SAM is the element-wise multiplication of the broadcasted final attention map and $X^e_{i-1}$.  At the same time, $X^d_i$ is accordingly passed through a convolutional layer and a nearest upsampling layer with a factor of 2 to recover the resolution. Then a Channel Attention Module (CAM) is used to concatenate and fuse the low-level feature from the SAM and the high-level feature from the upsampling operation. For the CAM, as in \cite{Hu_2018_CVPR},  the concatenated features are processed through the global average pooling layer, fully-connected (FC) layer with ReLU, and FC layer with sigmoid function to generate the channel attention map. Similar to the SAM, the output of the CAM is obtained by element-wise multiplication of the attention map and the original features. After the process of CAM, a convolutional layer is used to output $X^d_{i-1}$. Finally, a convolutional layer with sigmoid is required to process $X^d_{i-1}$ when necessary, and the sigmoid output $\sigma$ can be converted to the depth map $D_i$ with $1/(a\sigma+b)$, which is omitted in Fig. \ref{fig_VAB} for simplicity. Note that the SAM is removed from VAB when the internal feature from the encoder are unavailable.

\begin{table*}[t]
	%\vspace{10pt}
	\caption{Quantitative results on the Eigen split of KITTI dataset}
	\label{table_kitti}	
	\begin{center}
		\begin{threeparttable}
			\begin{tabular}{c|c|c||cccc|ccc|c}
				\hline
				\multirow{2}{*}{Method} & \multirow{2}{*}{Train}  &
				\multirow{2}{*}{Scale Factor}  & 
				\multicolumn{4}{c|}{The lower the better} &
				\multicolumn{3}{c|}{The higher the better} & \multirow{2}{*}{\#Param.}\\
				\cline{4-10}
				&&&Abs Rel & Sq Rel & RMSE & RMSE log & $\delta_1$ & $\delta_2$ & $\delta_3$&\\
				\hline
				SfMlearner \cite{zhou2017unsupervised}\dag &M& GT & 0.183 & 1.595 & 6.709 & 0.270 & 0.734 & 0.902 & 0.959 & 31.6M\\
				Monodepth2 \cite{godard2019digging} &M& GT & 0.115 & 0.903 & 4.863 & 0.193 & 0.877 & 0.959 & 0.981 & 14.3M\\
				DNet \cite{xue2020toward} &M& GT & 0.113 & 0.864 & 4.812 & 0.191 & 0.877 & 0.960 & 0.981& 14.3M\\
				SGDepth \cite{klingner2020self} &M+Sem&GT&0.113&0.835&4.693&0.191&0.879&0.961&0.981&14.3M\\
				Packnet-sfm \cite{guizilini20203d} &M& GT & 0.111 & 0.785 & 4.601 & 0.189 & 0.878 & 0.960 & 0.982&128M\\
				HR-Depth \cite{lyu2021hr} &M& GT & 0.109 & 0.792 & 4.632 & 0.185 & 0.884 & 0.962 & 0.983&14.6M\\
				Johnston et al. \cite{johnston2020self} &M& GT & 0.106 & 0.861 & 4.699 & 0.185 & 0.889 & 0.962 & 0.982&-\\
				Wang et al. \cite{wang2021can}&M&GT& 0.109&0.779&4.641&0.186&0.883&0.962&0.982&-\\
				FSRE-Depth \cite{jung2021fine}&M+Sem& GT & 0.105 & \textbf{0.722} & 4.547 & 0.182 & 0.886 & 0.964 & \textbf{0.984} &25.2M\\
				CADepth \cite{yan2021channel}&M& GT & 0.105 & 0.769 & \textbf{4.535} & \textbf{0.181} & \textbf{0.892} & 0.964 & 0.983&58.3M\\
				\textbf{VADepth (Ours)} w/o $L_{as}$ &M& GT & \textbf{0.104} & 0.774  &   4.552  &  \textbf{0.181} & \textbf{0.892}  &   \textbf{0.965}  &   0.983  &18.8M\\		
				\hline
				DNet\cite{xue2020toward} &M&  Cam. Height & 0.118 & 0.925 & 4.918 & 0.199 & 0.862 & 0.953 & 0.979 &14.3M\\
				Packnet-sfm \cite{guizilini20203d}&M+v& None & 0.111 & 0.829 & 4.788 & 0.199 & 0.864 & 0.954 & 0.980&128M\\
				Wagstaff et al. \cite{wagstaff2020self}&M& None & 0.123 & 0.996 & 5.253 & 0.213 & 0.840 & 0.947 & 0.978 &-\\
				Monodepth2 \cite{godard2019digging} with $L_{as}$ &M& None &   0.112  &   0.875  &   4.905  &   0.199  &   0.863  &   0.955  &   0.980 & 14.3M\\
				\textbf{VADepth (Ours)} with $L_{as}$ &M& None &  \textbf{0.109}  &   \textbf{0.785}  &   \textbf{4.624}  &   \textbf{0.190}  &   \textbf{0.875}  &   \textbf{0.960}  &   \textbf{0.982} & 18.8M\\
				\hline
			\end{tabular}
			\begin{tablenotes}
				\footnotesize
				\item All models are tested with the resolution of $192\times640$ unless otherwise specified, for the maximum depth of 80m. $\dag$ means the newer results from github with the resolution of $128\times416$. The best scores for each category are in \textbf{bold}. In the ``Train" column, we list the supervision for each method with M --- Self-supervised monocular supervision, Sem --- Semantic supervision, v --- velocity supervision. In the ``Scale Factor" column, ``GT" refers to determining a per-image scale factor using the ground truth depth  for median scaling based scale recovery, ``Cam. Height" indicates leveraging the known camera height w.r.t the ground surface to recover scaled depth, while ``None" means test without any post-processing. 
			\end{tablenotes}
		\end{threeparttable}
	\end{center}
	
\end{table*}

\subsection{Co-planar Points Detection on Ground Surface} 

The depth estimation model obtained by minimizing the baseline loss defined in (\ref{lbase}) can only directly output the relative depth. In order to train a depth estimation model that can output absolute depth maps, we need to utilize the geometric information that contains absolute scale. In most self-driving scenarios, the ubiquitous camera height can be considered as a known constant value. Therefore, the prior knowledge about camera height is suitable for constructing scale constraints by computing the camera height errors derived from the estimated depth.

To estimate the camera height from the predicted depth map, we need to detect the co-planar points on the ground surface. We observe that the middle-lower region of the image is part of the ground surface in almost all self-driving cases. Based on such geometric prior knowledge, we suppose that a small stationary rectangular region in image $I_t$ is the ground plane. We use two hyperparameters $\alpha_u=0.075$ and $\alpha_v=0.875$ to predefine an $H\times W$ binary mask $M_{rect}$ (as shown in Fig. \ref{fig_abs_loss}) to distinguish whether a pixel is in the rectangular region.  

\begin{equation} \label{prior_mask}
M_{rect}(u, v) = 
\begin{cases}
1 & \text{if } |0.5 - \frac{u}{W}| < \alpha_u \text{ and } \frac{v}{H} > \alpha_v,\\
0 & \text{others.}
\end{cases}
\end{equation}
Let $P_{rect}\in R^{N_{rect}\times 3}$ be a matrix for a set of 3D points $\{P_t(u, v)|M_p(u, v)=1\}$ in the predefined rectangular region, where $N_{rect}=2\alpha_uW(1-\alpha_v)H$ and $P_t(u, v)$ is calculated from the estimated depth according to \eqref{pt}. Based on these 3D points, we can fit a plane $Pl_{rect}$ that does not pass through the origin by solving the equation: $P_{rect}\boldsymbol n=\boldsymbol1$. We take the Moore-Penrose inverse of $P_{rect}$ to obtain the least-squares solution of the contradictory equations:
\begin{equation} \label{normal}
\boldsymbol n=P_{rect}^+\boldsymbol1.
\end{equation}
After plane fitting, we need to detect all 3D points co-planar with the plane $Pl_{rect}$ in the target view using a binary mask:
\begin{equation} \label{mask_p}
M_p(u, v) = 
\begin{cases}
1 & \text{if } |P_t(u, v)\boldsymbol n-1|<\delta,\\
0 & \text{others.}
\end{cases}
\end{equation}    
In $M_p$, an element whose value is equal to one indicates that the corresponding point is on the plane $Pl_{rect}$. The threshold $\delta$ is set to 0.01, which is used to judge whether the points are coplanar or not.

\subsection{Absolute Scale Loss} After co-planar points detection on the ground surface, we construct an absolute scale loss term using the camera height. To form a scale constraint, we need to compute the camera height from the estimated depth results: 
\begin{equation} \label{H}
H_{cam}(u, v) = P_t(u, v)\boldsymbol n_e,
\end{equation}
where ${\boldsymbol n_e}=\frac{\boldsymbol n}{||\boldsymbol n||_1}$ is the  normalized vector of $\boldsymbol n$. Using the estimated heights $H_{cam}$ and the ground truth height $h_{gt}$, we define the absolute scale loss as follows:
\begin{equation} \label{l_as}
L_{as}= \frac{1}{||M_p||_1}||M_p\circ(H_{cam}-h_{gt})||_1,
\end{equation}
where $\circ$ is Hadamard product operator. Combining $L_{as}$ with $L_{baseline}$, we have the final objective function:
\begin{equation} \label{L_total}
L = \frac{1}{S}\sum_{i=0}^{S-1}\mu L_{ph} + \lambda_{sm}L_{sm} + \lambda_{as}L_{as}.
\end{equation}
In our setting, $\lambda_{sm}$, $\lambda_{as}$ and $S$ are set to $0.001$, $0.01$ and $4$, respectively.

\section{EXPERIMENTAL RESULTS}
\begin{figure*}[t]
	\centering
	\includegraphics[scale=0.9]{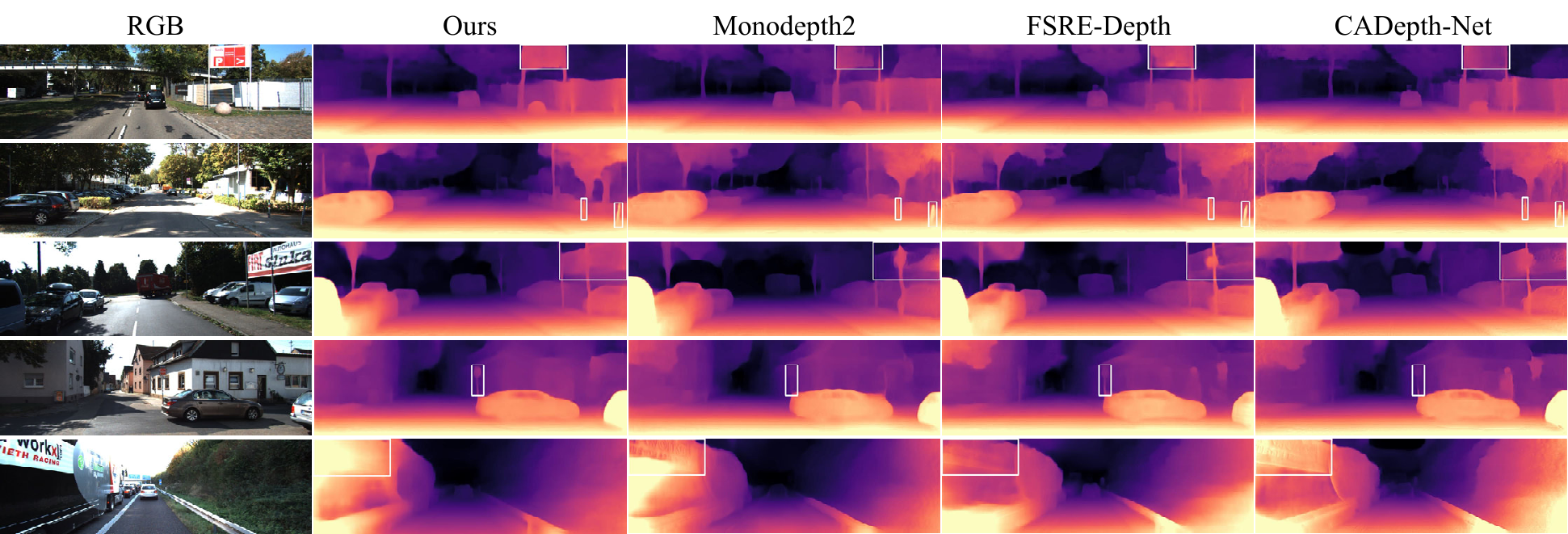}
	\caption{Quantitative results on the Eigen split of KITTI dataset. Compared to other methods \cite{godard2019digging, jung2021fine, yan2021channel}, our model not only outputs smoother depth in patterned planar regions, but also preserves better details for small objects. White boxes highlight the difference. Best viewed in color and zoom in.}
	\label{fig_kitti}
\end{figure*}

\begin{table*}[h]
	\caption{Generalization  performance on SeasonDepth}
	\label{table_SD}
	\begin{center}
		\begin{threeparttable}
			\begin{tabular}{|c|c|c|cc|cc|cc|cc|}
				\hline
				
				&&&&&\multicolumn{6}{|c|}{SeasonDepth test} \\ \cline{6-11}
				Method&Train&Resolution&\multicolumn{2}{|c|}{KITTI Eigen test}
				& \multicolumn{2}{|c|}{Average}
				& \multicolumn{2}{|c|}{Variance($10^{-2}$)}
				& \multicolumn{2}{|c|}{Relative Range} \\
				&&& $Abs Rel\downarrow$ & $a_1\uparrow$ & $Abs Rel\downarrow$ & $a_1\uparrow$ & $Abs Rel\downarrow$ & $a_1\downarrow$ & $Abs Rel\downarrow$ & $1-a_1\downarrow$\\
				\hline
				Eigen et al. \cite{eigen2014depth}\ddag & D & -& 0.203 & 0.702 &1.093 & 0.340 & 0.346 & \textbf{0.0170} & \textbf{0.206} & \textbf{0.0746}\\
				VNL(ResNext101) \cite{yin2019enforcing}\ddag & D & -& 0.072 & 0.938 &\textbf{0.306} & \textbf{0.527} & \textbf{0.126} & 0.166 & 0.400 & 0.290\\
				BTS(DenseNet161) \cite{lee2019big}\ddag & D & -& \textbf{0.060} & \textbf{0.955} &0.676 & 0.209 & 0.545 & 0.0650 & 0.405 & 0.129\\
				\hline
				SfMLearner \cite{zhou2017unsupervised}\ddag & M & $128\times416$&0.181 & 0.733 & 0.360 & 0.495 & 0.0801 & 0.0628 & 0.269 & 0.182\\
				Monodepth2 \cite{godard2019digging} & M & $192\times640$& 0.115 & 0.877 & 0.266 & 0.611 & 0.0410 & 0.0457 & 0.266 & 0.202\\
				FSRE-Depth \cite{jung2021fine}& M & $192\times640$& 0.105 & 0.886 & 0.256 & 0.624 & 0.0288 & 0.0283 & 0.227 & \textbf{0.158}\\
				CADepth-Net \cite{yan2021channel} & M & $192\times640$& 0.105 & 0.892 & 0.257 & 0.625 & 0.0447 & 0.0725 & 0.265 & 0.278\\
				\textbf{VADepth (Ours)} & M & $192\times640$& \textbf{0.104} & \textbf{0.892} & \textbf{0.230} & \textbf{0.667} & \textbf{0.0158} & \textbf{0.0215} & \textbf{0.205} & 0.179\\
				\hline
			\end{tabular}
			\begin{tablenotes}
				\footnotesize
				\item In the Train Column, D refers to ground truth depth supervision and M refers to self-supervised monocular supervision. $\downarrow$ means the lower the better while $\uparrow$ means the higher the better. Methods marked with $\ddag$ indicate that the corresponding test results are taken from \cite{hu2020seasondepth}.
			\end{tablenotes}
		\end{threeparttable}
	\end{center}
\end{table*}

\subsection{Implementation details}We use a single Nvidia GeForce RTX 2080 Ti to implement our models in Pytorch. In our setting, we use the VADepth network as our depth network with an input/output resolution of $192\times640$ . For pose estimation, we use a modified ResNet18 \cite{he2016deep} to predict a single 6-DoF ego-motion of two adjacent frames following \cite{godard2019digging}. The pose network adopts the same input resolution as the depth network. Both the encoder of the depth network and the encoder of the pose network are initialized with weights pretrained on ImageNet \cite{deng2009imagenet}. The depth network and pose network are jointly trained with Adam optimizer for 20 epochs. The learning rate is set to $5.0\times 10^{-5}$ for the first 15 epochs and then drops by half for the remaining epochs. The source code and models are available at \protect\href{https://github.com/xjixzz/vadepth-net}{https://github.com/xjixzz/vadepth-net}.

\subsection{Evaluation on KITTI}KITTI dataset \cite{geiger2012we} is the commonly used dataset for autonomous driving. We use the Eigen split \cite{eigen2014depth} of KITTI Stereo dataset to evaluate our method with the metrics proposed in \cite{eigen2014depth}. Following \cite{zhou2017unsupervised}, we also remove the static frames before training, which results in 39810 monocular triplets for training, 4424 images for validation and 697 images for test. The resolution of these images is  approximately $375\times1242$. All images are required to resize to $192\times 640$ for training and evaluation. At test time, the depth network only outputs the depth map with the resolution of $192\times 640$, which is then resized to the full resolution of the original RGB image for evaluation.

To evaluate the performance of our depth network, we first train VADepth network only with the baseline loss function. We report the results of applying median scaling at the top of Table \ref{table_kitti}, where we can see that our model achieves the best scores in the metrics of Abs Rel, RMSE log, $\delta_1$, and $\delta_2$, and ranks second or third in the remaining metrics. Among these methods, CADepth \cite{yan2021channel} achieves close performance to VADepth but requires much more parameters, due to the underutilization of visual attention, especially ignoring the channel and spatial interdependencies of low-level features. 

Besides, we also combine our absolute scale loss to train our model so that the depth network can directly output metrically accurate depth without post-processing that rely on the ground truth depth maps. As listed at the bottom of Table \ref{table_kitti}, we compare our methods with existing self-supervised absolute depth estimation methods that do not rely on the ground truth depth. Among these methods, our method achieves the best results on all metrics without requiring extra pretrained road segmentation network \cite{wagstaff2020self} or velocity supervision \cite{guizilini20203d} at training time, or post-processing \cite{xue2020toward} at test time.

The qualitative results are presented in Fig. \ref{fig_kitti}. Compared with \cite{godard2019digging, jung2021fine, yan2021channel}, our model predicts smoother and more accurate depth for patterned surfaces (e.g. the traffic sign in row 1, the banner in row 3, and the bus in row 5), which validates the effectiveness of our model to suppress unnecessary information. Meanwhile, our model also works well for thin things, such as the thin poles in row 2 and the traffic sign in row 4, which reflects the ability to highlight the important details. Taken together, these results illustrate that incorporating the attention mechanism throughout the whole depth network can help to improve the ability of the model to automatically distinguish important contextual information from redundant information for better depth estimation.

\subsection{Generalization performance}
SeasonDepth dataset \cite{hu2020seasondepth} is an MDE dataset that contains the same multi-traverse routes under 12 different environmental conditions. The test set of SeasonDepth contains 17225 RGB images and corresponding ground truth depth maps with the resolution of $768 \times 1024$. We use the model trained on KITTI to evaluate the generalization ability of our method on the test set of SeasonDepth without fine-tuning. Results in Table \ref{table_SD} show that our model generalizes better than existing supervised methods and self-supervised monocular training methods, which proves that our model is more capable of learning transferable feature representation.

\begin{figure}[h]
	\centering
	\includegraphics[scale=1.1]{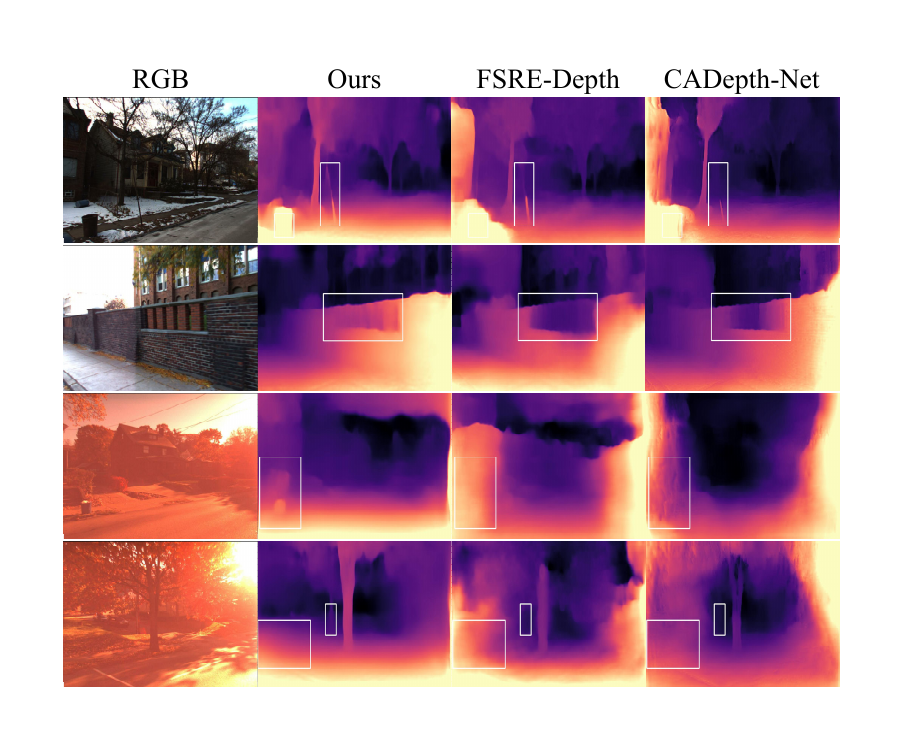}
	\caption{Quantitative results on the test set of SeasonDepth dataset. Our method generates depth maps that are more consistent with the scene structure in challenging environments.}
	\label{fig_season}
\end{figure}

We also provide the quantitative comparison in Fig. \ref{fig_season}. The inferior results of CADepth \cite{yan2021channel} and FSRE \cite{jung2021fine} suggest that applying channel attention in the decoder is not enough to obtain high-level features that are robust to lighting and weather changes. Despite extreme light conditions or challenging weather conditions, the output of our model still correctly reflects the scene structure. This further verifies that robust low-level feature representations extracted from our encoder and SAM are also important for forming robust high-level features.

\begin{table}[t]
	\caption{Ablation study of our network structure}
	\label{tab_vadpepth}
	\begin{center}
		\tabcolsep=0.1cm
		\begin{tabular}{c|c|cc||cccc}
			\hline
			\multirow{2}{*}{VAN} &
			\multirow{2}{*}{DAM} & 
			\multicolumn{2}{c||}{VAB} & 
			\multicolumn{4}{c}{The lower the better} \\
			\cline{3-8}
			&&SAM&CAM&AbsRel & SqRel & RMSE & RMSE log\\ 
			\hline
			&&&&	0.112  &   0.875  &  4.905  &   0.199 \\
			$\checkmark$&&&&   0.111  &   0.876  &   4.762  &   0.195  \\
			$\checkmark$&$\checkmark$ &&&   0.108  &   0.815  &   4.686  &   0.191  \\
			$\checkmark$&&$\checkmark$ & $\checkmark$ &   0.110  &   0.821  &   4.681  &   0.193\\
			$\checkmark$&$\checkmark$ & $\checkmark$ &&   \textbf{0.106}  &   0.789  &   4.672  &   \textbf{0.190} \\	
			$\checkmark$&$\checkmark$ && $\checkmark$ &   0.109  &   0.798  &   4.631  &   0.192\\
			$\checkmark$&$\checkmark$ & $\checkmark$ & $\checkmark$ &  0.109  &   \textbf{0.785}  &   \textbf{4.624}  &   \textbf{0.190}\\
			\hline

		\end{tabular}
		
	\end{center}
\end{table}

\subsection{Ablation Study}To verify the effectiveness of each component in our method, we perform an ablation study on the KITTI dataset.

Ablations of the VADepth network are reported in Table \ref{tab_vadpepth}, where all models are trained with the loss in (\ref{L_total}). We use Monodepth2 \cite{godard2019digging} as the baseline, which adopts a ResNet18-based encoder-decoder without any attention mechanisms. The model of row 2 is composed of a VAN \cite{guo2022visual} based encoder and a convolutional decoder similar to \cite{godard2019digging}, which performs better than the baseline. Adding either DAM or VAB leads to improvements on all metrics. Combining DAM with SAM or CAM can further improve the performance. Incorporating all modules together results in the best overall performance. 

We also ablate the absolute scale loss $L_{as}$ using different network architectures. It can be seen from Table \ref{table_ab_loss} that adding $L_{as}$ with proper $\alpha_u$ and $\alpha_v$ leads to the similar performance or even better performance compared to the baseline loss when ground truth depth maps are used for median scaling at test time. However, when testing without median scaling, the model trained without absolute scale loss performs poorly. When trained with absolute scale loss, both network architectures achieve good performance without any post-processing. These results indicate that the scale-aware depth network can be obtained with the camera height-based scale constraints. Besides, we also list the results of selecting different values for $\alpha_u$ and $\alpha_v$, which determine the size of the rectangular area defined in (\ref{prior_mask}), in the lower part of Table \ref{table_ab_loss}. Selecting a too large or too small rectangle degrades performance. Thus, we empirically select $\alpha_u$ and $\alpha_v$ as 0.075 and 0.875, respectively.

\begin{figure}[t]
	\centering
	\includegraphics[scale=0.95]{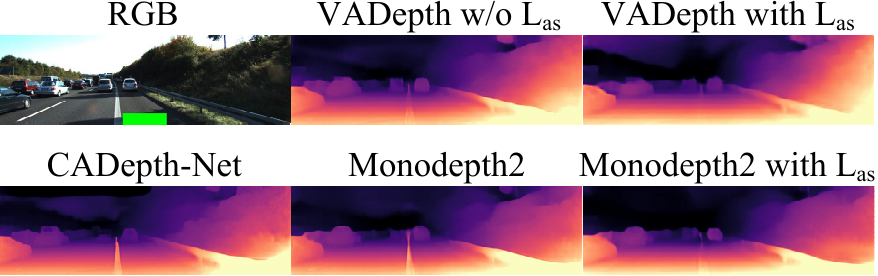}
	\caption{Quantitative results for ablation study of absolute scale loss on the KITTI dataset. The green box in the RGB image is the rectangular region defined by \eqref{prior_mask}. Using the absolute scale loss, the predicted results at the white solid line are smoother and more accurate.}
	\label{fig_abs_loss}
\end{figure}

Moreover, we find that absolute scale loss helps alleviate the impact of lane lines on depth estimation. As shown in Fig. \ref{fig_abs_loss}, although these networks can not predict smooth depth at the solid white line unless absolute scale loss is used, both VADepth and Monodepth2 \cite{godard2019digging} trained with the  absolute scale loss output more accurate depths at the lane line, which suggests that adopting supervision of camera height can lead to smooth depth estimates of the ground surface, in addition to absolute scaled depth. This further demonstrates that our network can effectively learn the geometric knowledge of autonomous driving scenarios with the proposed loss.

\begin{table}[t]
	%\vspace{10pt}
	\caption{Ablation study on absolute scale loss}
	\label{table_ab_loss}
	\tabcolsep=0.15cm	
	\begin{center}
		\begin{tabular}
			%{c|p{10pt}p{14pt}|cccc|c}
			{c|cc|cccc|c}
			\hline
			Scale&
			\multicolumn{2}{c|}{$L_{as}$}&
			%\multirow{2}{*}{$\alpha_u$, $\alpha_v$} &
			%\multirow{2}{*}{$\alpha_v$} &
			\multicolumn{4}{c|}{The lower the better} &
			\multirow{2}{*}{$\delta_1\uparrow$}\\
			\cline{2-7}
			Factor& $\alpha_u$ & $\alpha_v$ &AbsRel & SqRel & RMSE &RMSElog&\\%
			\hline
			\hline
			\multicolumn{8}{c}{Monodepth2}\\
			\hline
			\multirow{2}{*}{GT} & \multicolumn{2}{c|}{w/o $L_{as}$}  & 0.115 & 0.903 & 4.863 & 0.193& 0.877\\
			\cline{2-8}
			&0.075 & 0.875&   \textbf{0.112}  &   \textbf{0.864}  &   \textbf{4.804}  &   \textbf{0.190}  &   \textbf{0.878} \\
			
			\hline
			\multirow{2}{*}{None} & \multicolumn{2}{c|}{w/o $L_{as}$}  &   0.968  &  15.094  &  19.176  &   3.459  &   0.000\\
			\cline{2-8}
			&0.075 & 0.875&   \textbf{0.112}  &   \textbf{0.875}  &   \textbf{4.905}  &   \textbf{0.199}  &   \textbf{0.863} \\
			\hline
			\hline
			\multicolumn{8}{c}{VADepth}\\
			\hline
			\multirow{2}{*}{GT} & \multicolumn{2}{c|}{w/o $L_{as}$}& \textbf{0.104} & 0.774  &   4.552  &  0.181 & \textbf{0.892}\\
			\cline{2-8}
			&0.075 & 0.875&   0.105  &   \textbf{0.757}  &   \textbf{4.501}  &   \textbf{0.180}  &   0.891 \\
			
			\hline
			\multirow{10}{*}{None} & \multicolumn{2}{c|}{w/o $L_{as}$}  &   0.966  &  15.045  &  19.147  &   3.398  &   0.000\\
			\cline{2-8}
			&0.025 & 0.950 &     0.127  &   0.932  &   4.920  &   0.207  &   0.850     \\
			&0.050 & 0.900 &   0.110  &   0.870  &   4.719  &   0.191  &   0.876  \\
			&0.050 & 0.875 &   0.110  &   0.792  &   4.654  &   0.192  &   0.872     \\
			&0.075 & 0.900&   \textbf{0.109}  &   0.877  &   4.702  &   \textbf{0.190}  &   \textbf{0.879}      \\
			&0.075 & 0.875  &   \textbf{0.109}  &   \textbf{0.785}  &   \textbf{4.624}  &   \textbf{0.190}  &   0.875 \\
			&0.075 & 0.850  &  \textbf{0.109}  &   0.827  &   4.709  &   0.192  &   0.875     \\
			&0.100 & 0.875 &   \textbf{0.109}  &   0.825  &   4.680  &   \textbf{0.190}  &   0.876     \\
			&0.100 & 0.800 &   \textbf{0.109}  &   0.818  &   4.674  &   0.192  &   0.875    \\
			&0.200 & 0.600 &   0.126  &   0.863  &   4.858  &   0.209  &   0.850     \\			
			\hline
		\end{tabular}
	\end{center}
\end{table}

\section{CONCLUSION}

In this work, we presented a novel deep learning architecture for self-supervised monocular depth estimation and proposed an absolute scale loss to supervise the scale of estimated depth maps. Experimental results show that our method achieves state-of-the-art performance with or without median scaling on the KITTI dataset and generalizes well on the unseen changing environments. In the future, we plan to further improve the generalization capacity and interpretability of the depth estimation network by visualizing the attention map and using the results of semantic segmentation to guide the network to strengthen the responses of different channels to specific semantic information or improve the distinction between important features and noises.

\ifCLASSOPTIONcaptionsoff
  \newpage
\fi

\end{document}